# Conjoined Predication and Scalar Implicature

Ratna Kandala

n038k926@ku.edu

Department of Psychology,

University of Kansas, USA

**Abstract**

Magri (2016) has discussed two puzzles raised by conjunction. While the first puzzle has not been resolved, a solution to the second puzzle has been proposed by Magri. The first puzzle conceals an interrelationship between quantification, collective/concurrent interpretation, and contextual updates, the aspects of which have not been explored. In brief, the puzzle is that certain variants of sentences such as #*Some Italians come from a warm country* involving conjunction as in #*(Only) Some Italians come from a warm country and are blond* remain odd despite the fact that no alternative seems to trigger the mismatching scalar implicature. In this paper, we offer a conceptual analysis of Magri's first puzzle, by first presenting it in the context of the theory in which it arises. This paper proposes that the oddness arises due to the collective-concurrent interpretation of the conjunctive predicate, as underlined in #*(Only) Some Italians come from a warm country and are blond* that ends up giving rise to an indirect contextual contradiction. It is suggested that the generation of scalar implicatures may have pragmatically governed facets not fully conditioned by accounts of exhaustification-based grammatical licensing of scalar implicatures.

## 1. Introduction

Magri (2016) has discussed two puzzles raised by conjunction which we discuss in brief. Before looking into the first puzzle, we shall briefly discuss the second puzzle. The second puzzle begins with the observation that a sentence such as (1) triggers a scalar implicature (SI) that sentence (2) is false (or alternatively, the negation of (2) is true). Additionally, sentence (1) generates (2) as a scalar alternative (A) and, given its SI that (2)/A is false, gives rise to a conflict with common knowledge that all Italians actually come from a warm country[1].

(1) *Some Italians come from a warm country.*
   **SI**: ¬ All Italians come from a warm country.
   **A:** All Italians come from a warm country.
(2) *All Italians come from a warm country.*

If this alternative (A) is not generated and consequently its negated version is not generated as an SI of (1), no mismatch with common knowledge can take place. This is the exact reason why Magri (2016, pp. 2) says "A crucial tenet of this theory is that in all these cases of oddness we ought to be able to pinpoint **an alternative which suffices to trigger the mismatch with common knowledge** and which is furthermore justified independently, namely through the scalar behavior of formally analogous but felicitous sentences." (emphasis bold-marked)

It is to be noted that a distinction is thus made between a scalar implicature and an alternative[2], given that not all alternatives go on to make SIs (see Horn, 1972, 2000; Fox, 2007). Further, if the meaning of (1) is strengthened with 'only', as in (3) below, this ends up generating a contextual contradiction (a contradiction that is based on **contextual** rather than logical inconsistencies with (1) because (3) conflicts with the common knowledge that all Italians come from a warm country, and both (3) and this piece of common knowledge cannot be contextually valid at once).

(3) *Only some Italians come from a warm country.*

Since (2) is not part of the truth-conditional meaning of (1) (when strengthened with 'only', as in (3)) and has scale-based meaning contrast in terms of Horn scales ('a few', 'some', 'many',

---

[1] An anonymous reviewer has pointed out that 'some' in sentence (1) need not mean 'not all', for after all it may be underinformative and mean 'Some (and possibly all) Italians come from a warm country' (see also Bott & Noveck, 2004). This may hold only if the theory of scalar implicature generation via the negation of alternatives is not presupposed. As Magri (2009, 2011) assumes that scalar implicatures are generated blindly without taking contextual common knowledge into account, this presupposition leads to the very puzzle here within the context of the theory concerned.

[2] The scalar implicature is what is implicated through the use of an utterance containing a scalar expression (such as 'all', 'some', 'most' etc.) but not overtly said (see Grice, 1975; Levinson, 2000), whereas an alternative is to be understood as a sentence constructed with an alternative quantifier (or even modal) expression having a difference on a scale of quantity or degree (such as the scale < not all, few, none> or <sometimes, often, always>) (see also Sauerland, 2004).

'most', 'all') with respect to (1), sentence (2) can act as a scalar alternative[3]. The underlying idea is that the stronger words from a scale (such as 'all') are needed for the interpretation of weaker words from the same scale (such as 'some') whenever an implicature is derived, but the stronger words can be interpreted without any reference to the weaker words. That is mainly because stronger words on the scale are more informative than weaker words on the scale, and hence less informative words from the scale can be made sense of in terms of the stronger words. However, this does not necessarily hold in neutral contexts where partial information is available to speakers, as in *Some of today's gifts are clothes*, where the speakers may not have checked all the gifts by opening them and hence 'some' may not imply 'not all' for listeners. But, if number words are used in such contexts, as in *2 of today's gifts are clothes*, 'some' strongly implies 'not all' for listeners (see Zhang et al., 2023).

To put things in context, Magri's theory of oddness in scalar implicature (2009, 2011) postulates that an existentially quantified sentence triggers the SI that its universally quantified version is false. But common knowledge dictates that if the existentially quantified sentence holds true, the its universally quantified version must also be true. The oddness arises between the SI generated and the entailment from common knowledge. The crucial steps can be specified in the following manner (see Magri, 2009, pp. 3).

(i) Due to the presence of existential quantification, (1) triggers the SI that not all Italians come from a warm country.
(ii) But common knowledge entails that if some Italians come from a warm country, then all of them do, because they come from the same country.
(iii) The oddness of (1) thus follows from the mismatch between (i) and the common knowledge (ii).

Crucially, the generation of SI in Magri's theory of oddness is blind to common knowledge from the context at an initial step, because step (i) specifies the generation of the relevant SI while it is at step (ii) that context comes into the picture. Magri (2011) argues that it is this blindness that explains the oddness of a sentence with an individual-level predicate such as 'tall', as in *#John is sometimes tall*. Unless the SI ¬*John is always tall* is generated without consideration of, or access to, the common knowledge that a person's tallness is a permanent property, it is hard to see how *#John is sometimes tall* can be judged to be odd. Another important notion within this account is the notion of relevance. When the alternative sentences are considered for the generation of an SI via negation, only those alternatives that are relevant in a given context are factored in.

---

[3] But one way of checking that this is so is to appeal to the defeasibility of (scalar) implicatures. Since scalar implicatures (in fact, implicatures in general in Gricean theory) are defeasible (see Levinson, 1983; Asher, 2012), when the truth of sentence (1) with 'some' is challenged, the implicature sentence (1) generates may be cancelled and sentence (2) can then act as an alternative statement for (1).

Interestingly, the idea that SI effects can arise via the generation of alternatives that are obtained by substitutions of quantifiers into sentences as part of grammatical mechanisms, however, contrasts with the view adopted in Relevance Theory (Sperber & Wilson, 1995; Wilson & Sperber, 2004). That is because in Relevance Theory SI effects are usually regarded as the result of contextual enrichment of logical forms generated by grammar as a post-grammar process. In any case, it is evident that Magri (2016, pp. 15-16) has notably drawn upon the account of alternatives described in Sauerland (2004), which is actually grounded in the work of Horn (1972, 1989). But, interestingly, when two predicates are conjoined in the scope of the existential quantifier 'some', as in (4) below, (4a) (that is, 'All Italians come from a warm country and are blond'), although otherwise independently justified as an alternative of (4), does not *appear* to act as an alternative of (4), crucially because the negation of (4a) is consistent with the common knowledge that all Italians come from the same country (by disregarding any coincidental connection between nationality and blondness)[4]. This makes it harder to see how the negated version of (4a) can act as an SI of sentence (4), given that (4a) does not seem to act as an alternative of sentence (4) *and also* that a scalar implicature is generated by way of the negation of an alternative (Horn, 1989, pp. 232). The point here is that the negated version of (4a) as the SI of (4) is supposed to eliminate the oddness of (4) in virtue of triggering no mismatch with common knowledge (that all Italians come from the same country). Nonetheless, (4) is odd.

(4) *(Only) Some Italians come from a warm country and are blond*.
    a. All Italians come from a warm country and are blond.
    \* **A:** All Italians come from a warm country.

---

[4] It is noteworthy that Anvari's (2018, pp. 714, 719) principle of Logical Integrity also cannot help explain the oddness in (4). According to Logical Integrity, if a sentence does not *logically* entail, but contextually entails, one of its alternatives in a context, then in that context the sentence would be infelicitous. In a context where the common knowledge is that all Italians come from a warm country, sentence (1) certainly does not logically entail its alternative sentence (2), but contextually entails it, because if some Italians come from a warm country, all of them do too. Interestingly, this strategy appears to fail for (4) because an alternative of it does not seem to be available. Since the condition for Logical Integrity is not met, it is not clear how it can apply here to account for the oddness in (4).

That is why Magri (2016) says "(23) a. #Some Italians come from a warm country. b. #Some Italians come from a warm country and are blond. As no mismatching alternative seems readily available, the oddness of sentence (23b) challenges the very idea of a theory of oddness based on mismatching scalar implicatures. I have left this first puzzle unsolved." (pp. 13-14)

As a matter of fact, this point can also be appraised better with the help of another parallel example provided by Pistoia-Reda and Romoli (2017, pp. 57). Given the analogous sentence *Some men are mortal and tall*, the negation of *All men are mortal and tall* does not conflict with the common knowledge that all men are mortal and, instead, it is consistent with that piece of common knowledge. It may be further noted that (4) does not give rise to (2) as an alternative either (as indicated by the asterisk before *A: All Italians come from a warm country* in (4)), and hence this does not lead to any logical or contextual contradiction. Nonetheless, (4) remains odd. This account of oddness is grounded in the grammatical theory of compositional calculation of implicatures (Chierchia, Fox & Spector, 2011) within which the relevant SI is generated in blindness to the common knowledge in the first place only to conflict with common knowledge, even though on a Gricean account (Grice, 1975) implicatures can be canceled anyway and it is not clear why the infelicity of (1) or even of (4) cannot be blocked by appealing to a pragmatic explanation along the line of a Gricean account (Pistoia-Reda & Romoli, 2017; see also, Schlenker, 2012). There are perhaps other matters involved that are responsible for the generation of this SI-caused-oddness effect. On the one hand, something over and above a Gricean pragmatic explanation may be involved when the SI-caused-oddness effect arises for (1). On the other hand, the relevant alternative for (3)/(4) does not emerge and hence it appears that the SI-caused-oddness effect does not apply to (3)/(4) as straightforwardly as it does to (1).

Significantly, Magri (2016) argues that the oddness in the case of (3)/(4) disappears especially when the relevant sentence is preceded by a sentence that leads to a contextual manipulation, as shown below. This engenders the second puzzle.

(5) *Italians come from a warm country. Some of them, come from a warm country and are blond.*

(6) *Italians come from a warm country. Only some of them, come from a warm country and are blond.*

This happens, as Magri argues, because in the case of (1) the generation of the SI for (1) is not sensitive to contextual manipulations but to the relevance of the alternatives. Since (2) remains relevant when the sentence *Some Italians come from a warm country (and are blond)* is uttered, contextual contradictions do arise. Here the question *Do Italians come from a warm country?* (plus additional questions concerning their blondness) remains entrenched. But when the preceding discourse, as in (5)-(6), settles this question at the outset, (2) no longer remains relevant. This is Magri's solution to the second puzzle. However, the fact that (4) does not give rise to (4a) or (2) as a scalar alternative (despite the relevance of (2)) and hence is (incorrectly) supposed to sound fine remains puzzling. While there exists a mismatching alternative in the case of (1), which is definitely odd, the conjoined version in (4) has no mismatching alternative that can lead to a conflict with common knowledge and *yet* is odd. This is exactly the first puzzle.

To take stock, the first puzzle arises primarily because of the absence of a mismatching alternative. Let's first consider the case in (1). Sentence (1) triggers its SI due to the contrast of Horn scales vis-à-vis (2). This inference/implicature is in some way part of the truth-conditional meaning of (3), which is a variant of (1)[5]. That is because if (only) some Italians come from a warm country, it must be the case that not all do. Hence, the negated version of (2) remains preserved as an implicature unless it is canceled. Therefore, (2) must count as a scalar alternative of (1). This is (again) substantiated by the fact that (3) can be felicitously rebuffed by saying *You are wrong: all Italians come from a warm country*, and cannot be continued in the discourse context with #*And all Italians come from a warm country*. Now, in contrast, we may consider the case of (4) with 'only' placed before 'some'. We can see how the parallelism with (1) or (3) breaks down in the case of (4). It may be noticed that (4) cannot be felicitously rebuffed by saying *#You are wrong: All Italians come from a warm country*, and yet can be continued in the discourse context with the sentence *And all Italians come from a warm country*. The puzzle here consists in the presence of oddness in (4) even when the negation of the alternative (4a) does not give rise to any oddness. So, if the negation of the alternative (4a) does not mismatch or conflict with common knowledge, no oddness should arise, but sentence (4) still remains odd. The whole pattern of oddness in (4) is schematized in Figure 1.

---

[5] Perhaps the only way in which the SI of (1) can be part of the truth-conditional meaning of (3) is via *exhaustification* (to be discussed in section 2.2), irrespective of whether it is done in a syntactically targeted feature-checking process (Chierchia, 2013) or in a grammatically determined manner suitably restricted by a contextual variable (Fox, 2007).

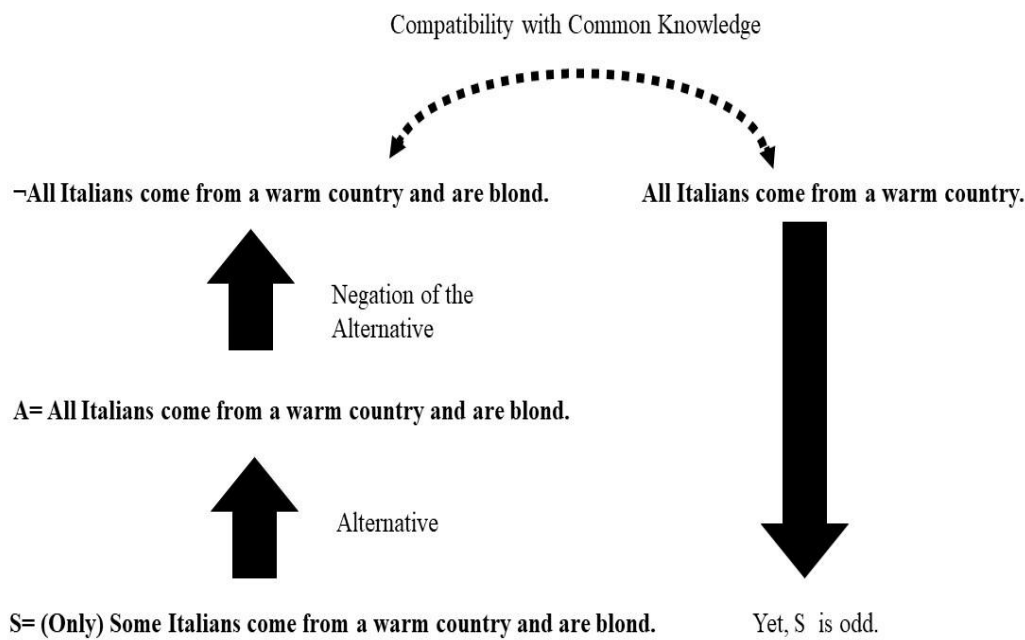

Figure 1: The graphical representation of the pattern of oddness in (4)

This paper is structured as follows. Section 2 goes on to offer a plausible pragmatic solution to the puzzle which is grounded in the explication of the collective/concurrent interpretation of the predicate conjuncts in (4). Also, this section reflects on the theoretical relevance and implications of the analysis. Additional evidence for the collective/concurrent interpretation of the predicate conjuncts in (4) is adduced in section 3 by showing that a distributive reading of the conjuncts does not lead to any oddness in (4). Finally, some concluding remarks are offered in section 4.

**2. Collective interpretation and the role of indirect contextual contradiction**

One way of approaching the puzzle is by making reference to the point that the collective-concurrent interpretation of the two predicate conjuncts in (4) ends up giving rise to a contextual contradiction, albeit indirect.

*2.1 The collective interpretation of the predicate conjuncts*
In order to flesh out the account proposed, we may first establish that the predicate conjuncts in (4) have a collective interpretation. Hence, we may designate the relevant sets.

(7) A=the set of Italians

B'=the set of those individuals that come from a warm country

B"=the set of those individuals that are blond

The sentence in (4) indicates that $A \cap B' \cap B'' \neq \phi$. The quantifier 'some' is intersective in virtue of being **conservative** on both predicates *come from a warm country* and *are blond* on the one hand and also on 'Italians' as a predicate/set of individuals on the other hand (Keenan & Stavi, 1986; Keenan, 2006). This would license the conclusion formulated in (8).

(8) **some** (A) (B) = **some** (A) (B'$\cap$B") =1 *if and only if*

$A \cap B \neq \phi$

$\equiv A \cap (A \cap (B' \cap B'')) \neq \phi$                         [conservative on set A]

$\equiv (A \cap (B' \cap B'')) \cap (B' \cap B'') \neq \phi$               [conservative on B= B'$\cap$B"]

$\equiv A \cap B' \cap B'' \neq \phi.$                                ['$\equiv$' may be read as 'that is']

Thus, the conjoined predicate *come from a warm country and are blond* forms a kind of restriction for the existential quantifier 'some'. Under this restriction, B' is restricted to those members that lie in B" and vice versa. Now what appears significant is that the (restricted) set of Italians that come from a warm country *and* are also blond are part (say, an improper subset) of a number (or group) of Italians who come from a warm country. Hence, if it is true that some Italians come from a warm country *and* are blond (that is, if **some** (A) (B'$\cap$B") is true), it follows that these Italians are part of a(n) (indefinite) number (or group) of Italians that come from a warm country. Let this group be specified by means of a special quantifier symbol $Q^i$ (A) (B'). There is nothing special about $Q^i$, except for the fact that we do not have any natural language quantifier for the (indefinite) number we are interested in. But for the sake of simplicity, we may assume that $Q^i$ (A) (B') = A(n) (indefinite) number of A$x$ is B'$x$. With this formulation, we may now recognize that if $Q^i$ (A) (B') is conservative (on A as the first predicate argument), then (9) must hold.

(9) $Q^i$ (A) (B') =1 *if and only if*

$Q^i$ (A) (B') = $Q^i$ (A) (A$\cap$ B')

Crucially, it is now important to notice that (4) entails (designated by ⊩) $Q^i$ (A) (B'), as stated in (10) with its natural language counterpart in (11).

(10) **some** (A) (B'∩B'') ⊩ $Q^i$ (A) (B')

(11) *Some Italians come from a warm country and are blond* **entails** *A(n) (indefinite) number of Ax is B'x.*

This entailment pattern is shown in Figure 2. We shall now discuss the role of exhaustification in (4).

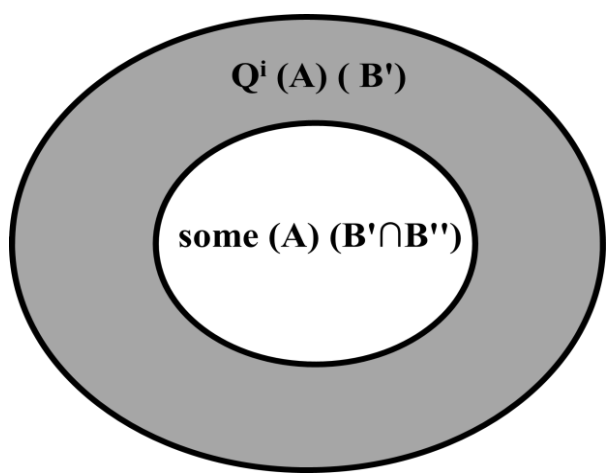

Figure 2: The entailment relation between **some** (A) (B'∩B'') and $Q^i$ (A) (B')

## *2.2 The role of exhaustification and presupposition*

Now, the first thing we can check is whether *exhaustification* is actually implemented for (4). Exhaustification is a kind of inference that establishes an expression as a scalar implicature via strengthening, and thus it limits the possibilities of various implicated meanings for a given sentence/expression (Chierchia, Fox & Spector, 2011). Thus, for instance, the very implicature that *not all* Italians come from a warm country is obtained from (1) via exhaustification. We note that the exhaustification of the scalar (structural) alternatives in (4) is *not* vacuous. The exhaustification of a scalar alternative can be vacuous when the negation of the alternative can be in conflict, or contradict, the original sentence whose SI is supposed to be generated. For instance, *John thinks that SOME of the students smoke* with its prosodic prominence on 'some' generates the SI that John does not think that all of the students smoke, whereas a sentence with a decreasing context such as *John doubts that SOME of the* *students smoke* seems infelicitous on a view of SI generation based on the negation of an alternative, chiefly because the SI that John does not doubt that all of the students smoke is in conflict with *John doubts that SOME of the students smoke* (Spector & Sudo, 2017: 490). That is because the negation of the alternative of (4), that is, the negation of *All Italians come from a warm country and are blond*, is consistent with (4) and also with the common knowledge that all Italians come from the same country, as also pointed out by Magri (2016, pp. 4). Nonetheless, the oddness persists. However, the alternative of (4)—that is, (4a)—has a presupposition that entails, and is thus stronger than, the presupposition of (4). This is shown in (12) below.

(12) **all** (A) (B'∩B")
PRESUPPOSITION 1: There are Italians and all of them come from a warm country and are blond.
**some** (A) (B'∩B")
PRESUPPOSITION 2: There are Italians and some of them, but not all, come from a warm country and are blond.

Clearly, PRESUPPOSITION 1 is stronger than PRESUPPOSITION 2, and in being so, PRESUPPOSITION 1 does not include common knowledge. A stronger statement is not supposed to be common knowledge and hence excludes common knowledge since it tends to be more informative, whereas common knowledge is something that is known and hence less informative. However, we know that all Italians come from a warm country. Hence, PRESUPPOSITION 1 is in conflict with the contextually available common knowledge. This follows from the *Presupposed Ignorance Principle* (Spector & Sudo, 2017, pp. 494), which dictates that if the presupposition of an alternative (A) of a sentence (S) is stronger than the presupposition of the sentence (S) itself, infelicity ensues when the stronger presupposition, in being strong, is not supposed to be commonly known *and yet* it is actually common knowledge. This also accords with a similar line of reasoning/understanding that SIs are presuppositions and oddness ensues from their conflicts with common knowledge (see Bassi, Del Pinal & Sauerland, 2021, pp. 29-36). From a related perspective, Del Pinal (2021) has recently proposed that even if the presupposition and the assertive content of a sentence are each consistent with the common ground, the update of the common ground with the presupposition can be inconsistent with the assertive content of the sentence[6], thereby giving rise to oddness. It is easy to notice that the oddness is not accounted for in this way, precisely because the common ground containing the statement that all Italians come from a warm country, along with the update via the inclusion of PRESUPPOSITION 2, is not also going to be inconsistent with the assertive content of (4).

In any event, even though the *Presupposed Ignorance Principle* explains the infelicity of (4), as well as of (1) along similar lines, the difference in oddness is not accounted for by this principle. Crucially, the exhaustification of the *all*-alternative of (1) is in conflict with both the common knowledge that all Italians come from a warm country and the presupposition of the *all*-alternative of (1). In contrast, the exhaustification of the *all*-alternative of (4) is *only* in conflict with PRESUPPOSITION 1, *but not* with the common knowledge that all Italians come from a warm country. Now, this needs further probing, and hence we turn to the entailment: **some** (A) (B'∩B'') ⊩ $Q^i$ (A) (B').

## *2.3 The emergence of an indirect contextual contradiction*

Interestingly, we note that if **some** (A) (B'∩B'') ⊩ $Q^i$ (A) (B'), then it follows that PRESUPPOSITION 2 ⊩ $Q^i$ (A) (B'). Therefore, by transitivity, PRESUPPOSITION 1 ⊩ $Q^i$ (A) (B'). Since the quantity of $Q^i$ is indefinite, $Q^i$ can take values from different scalar-alternative quantificational expressions that are *at most as complex* as **some** in (4), along the line of thinking in Katzir (2007), Fox and Katzir (2011). Thus, for instance, 'all', 'most', 'many', etc. are as complex as 'some'. So, $Q^i$ → {all, most, …}, and hence $Q^i$ (A) (B') can have a disjunctive form: [**all** (A) (B') ∨ **most** (A) (B') ∨ **…** ∨ $\mathbf{Q_k}$(A) (B')], where $\mathbf{Q_k}$(A) (B') is another plausible alternative compatible with $Q^i$ (A) (B'). Therefore, PRESUPPOSITION 2 ⊩ [**all** (A) (B') ∨ **most** (A) (B') ∨ **…** ∨ $\mathbf{Q_k}$(A) (B')].

Now, if we make a distinction between primary implicatures (that is, negated certain propositions) and secondary implicatures (certain negated propositions) the way Sauerland (2004) has done, the primary implicature that ¬ **K all** (A) (B') is not generated at all for (4) (here, **K** represents a certainty operator). That is so because from common knowledge we know that **K all** (A) (B') and, at the same time, it also holds that **P all** (A) (B') (here, **P** represents a possibility operator), since it will possibly be true that **all** (A) (B'), given [**all** (A) (B') ∨ **most** (A) (B') ∨ **…** ∨ $\mathbf{Q_k}$(A) (B')], as dictated by the logic of disjunction. Then why does the oddness persist in (4)? The reason is that ¬ **K all** (A) (B') follows from the disjunction structure of [**all** (A) (B') ∨ **most** (A) (B') ∨ **…** ∨ **Q**$_k$(A) (B')] which is itself entailed not only by both our common knowledge (that **all** (A) (B') is true) and **some** (A) (B'∩B") but also by both PRESUPPOSITION 1 and PRESUPPOSITION 2. Further, the secondary implicature that **K** ¬**all** (A) (B') can also arise because this is fully consistent (in Sauerland's sense) with both **some** (A) (B'∩B") and ¬ **K all** (A) (B'). In contrast, the primary implicature that ¬ **K all** (A) (B') is generated from (1), but our common knowledge dictates that **K all** (A) (B'). This is what gives rise to the contextual contradiction, and also blocks the secondary implicature that **K** ¬**all** (A) (B'). The overall scenario is sketched out in Figure 3 below.

---

[6] Such arguments are also related to strategies that attempt to deal with scalar implicatures at the level of presupposition (see Marty & Romoli, 2022).

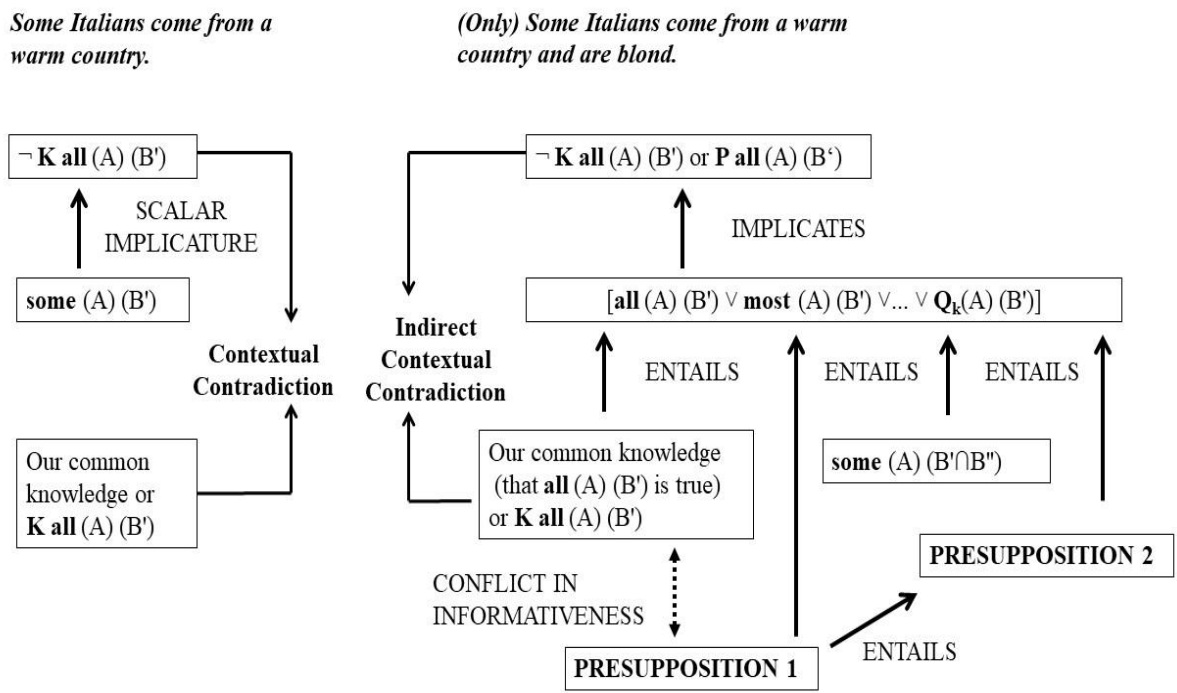

Figure 3: The overall pattern of difference in oddness between (1) and (4)

It is noteworthy that the exhaustification of the *all*-alternative of (4), that is, the output of the operation of exhaustification— ¬*All Italians come from a warm country and are blond*—is not in conflict with the common knowledge that **all** (A) (B'), that is, with **K all** (A) (B'). But, given that **some** (A) (B'∩B"), containing the conjoined predicate (B'∩B"), gives rise to [**all** (A) (B') ∨ **most** (A) (B') ∨ **…** ∨ **Q**$_k$(A) (B')] and also entails it, [**all** (A) (B') ∨ **most** (A) (B') ∨

**… ∨ Q<sub>k</sub>(A) (B')**] ends up implicating ¬ **K all** (A) (B')/**P all** (A) (B'). It is this implication of ¬ **K all** (A) (B')/**P all** (A) (B') that leads to a conflict or an *indirect contextual contradiction* with the common knowledge that **all** (A) (B')/**K all** (A) (B'). The implication of ¬ **K all** (A) (B')/**P all** (A) (B') from, [**all** (A) (B') ∨ **most** (A) (B') ∨ **…** ∨ **Q<sub>k</sub>**(A) (B')] is not, of course, automatic, but based on contextual updates.

This suggests that it is not the exhaustification of scalar alternatives but conjoined predication with its collective-concurrent reading of events that appears to generate SI-like effects even when the exhaustification of scalar alternatives is not vacuous. This phenomenon may be called a *scalar implicature hole*, because the usual exhaustification-based scalar implicature becomes weak or diminished and the implications generated by the set of entailments of a sentence can give rise to implicatures that go on to become entrenched by virtue of being not only structurally constrained (structured by conjoined predication) but also dependent on the dynamics of discourse governed by pragmatic constraints. Therefore, the theoretical implication is that Margi's account of mismatching alternatives and the grammatical view of SI it is based on, as in Horn (1972), Sauerland (2004), Katzir (2007), Fox and Katzir (2011), are incomplete. That is because the generation of SI via alternatives algorithmically generated through substitution into sentences only *partly* takes into consideration effects of context modeled as a restrictive mechanism. Relevance plays a significant role in this mechanism. This is akin to context insensitivity or opacity (see also Schlenker, 2012) which is what we observe especially in (4). However, it may be noted that this insensitivity or opacity can be (partially) restrictive in nature in order for the relevance of alternatives to figure in the account of scalar possibilities, (Pistoia-Reda, 2017).

The emergence of SI-like effects via the implications generated by the set of entailments of a sentence warrants a greater degree of contextual sensitivity. This conclusion is in line with experimental studies on the contextual sensitivity of SI (Zhang et al., 2023). Likewise, this kind of emergence of SI-like effects cannot be easily accounted for within Relevance Theory, for SI is supposed to be a matter of contextual enrichment in a post-grammatical process within Relevance Theory, whereas the implications generated by the set of entailments of a sentence partly ride on the grammar of conjoined predication in the present case.

In any case, this helps such implicatures to be patterned on effects that are partly due to grammatical licensing and partly due to the (locally) contextual nature of implications that are not overtly expressed but need to be deciphered from the intentions of speakers and also *questions under discussion* (Roberts, 1996; Burton-Roberts, 2010; Mayol & Castroviejo, 2013; Geurts, 2009). It seems that in the case of (4) the question under discussion (a matter of a relevant question for which (4) is uttered) does not change (the question *Do Italians come from a warm country and are they blond?*) and the utterance of (4) is intended, given that the more manifest a speaker's intention behind an utterance is, the harder it is to cancel it[7] (see Burton-Roberts, 2010, pp. 151). The example in (13) shows this very clearly.

(13) Speaker A: *Do Italians come from a warm country and are they blond?*

Speaker B: *Some Italians come from a warm country and are blond. And (in fact) all Italians come from a warm country.*

Speaker B: *Some Italians come from a warm country and are blond. #And (in fact) not all Italians come from a warm country/#And (in fact) it is not that all Italians come from a warm country*.

While, in response to Speaker A, Speaker B can continue the conversation by uttering *And (in fact) all Italians come from a warm country*, this does not seem to be possible if Speaker B continues the conversation with *And (in fact) not all Italians come from a warm country*. That is so because the question under discussion has not changed, and the utterance *And (in fact) not all Italians come from a warm country* or *And (in fact) it is not that all Italians come from a warm country* is a response to a different question, that is, *Do all Italians come from a warm country?*, which is definitely not at-issue. Thus, the implication that all Italians come from a warm country remains entrenched.

## 3. Further evidence: The distributive interpretation is blocked

This section provides further evidence that it is the collective interpretation[8], *but not* the distributive interpretation, which mainly contributes to the oddness in the first puzzle. In

---

[7] In the case of a conversation like the following, the speaker (that is, Ann) cannot simply cancel the implicature that Charles is a plagiarist without contradicting what she intended to say, mainly because the speaker's intention is quite manifest here.

Max: Do you ever speak to Charles?
Ann: I never speak to plagiarists.

[8] There is also independent psycholinguistic evidence that people tend to find sentences such as *The boys are sitting and reading* unacceptable half of the time on a split-situation reading (Poortman, 2017). This collective interpretation of compatible predicates can be extended to B' and B'' in the current context.

order to show this, we make a slight modification to (4) to make the distributive interpretation most favorable. The distributive reading of the two predicate conjuncts arises from (14). Here those Italians coming from a warm country and those that are blond need not be the same.

(14) *Some Italians come from a warm country and some Italians are blond*.

Thus, one can rebuff it saying *You are wrong: All Italians come from a warm country*. Besides, if the conjunction 'and' is not included under the scope of 'some', one can rewrite the odd sentence as (15).

(15) *Italians come from a warm country and (only) some of them are blond.*

Then (15) just like (1) triggers the scalar implicature ¬**all** (A) (B''). This approach works in a restricted domain of Italians, though. In any case, this demonstrates that the distributive reading does not give rise to the idiosyncratic situation (4) does.

Interestingly, Pistoia-Reda and Romoli (2017) suggest that the relevance of two conjuncts in a sentence can be derived from that of the whole conjoined sentence. In this regard, the *relatedness* of the conjuncts makes a crucial difference. The following pair of examples shows this well.

(16) *#Some of the Portugal players have won Euro 2016 and are tall*.

(17) *Some of the Portugal players have won Euro 2016 and left the team*.

Pistoia-Reda and Romoli (2017) argue that (17) does not give rise to any oddness while (16) does in view of the common knowledge that *all* players of a country (Portugal, in this case) win[9]. So, any account of the oddness in (1) and also (4) has to account for the absence of any oddness in (17) as well. It needs to be observed that the conjuncts in (4) have a collective and concurrent-situation reading but (17) does not seem to have the concurrent-situation reading, even though it appears that it can have the collective interpretation because those Portugal players who won Euro 2016 are exactly the individuals that also left the team. Hence the pragmatic conception of relatedness Pistoia-Reda and Romoli refer to can be cashed out in terms of the concept of event ordering/sequencing. That is because the events of the Portugal players having won Euro 2016 and their leaving the team must happen in a sequence, for the

---

[9] Part of the reason why (16) is odd but (17) is not is arguably the hearer's reaction to the mismatch in tenses or times (have won vs. are tall).

Portugal players cannot leave the team first and then win Euro 2016. We have already noted in section 2 that **some** (A) (B) = **some** (A) (B'∩B'') =1 iff A∩B ≠ ϕ holds for (4). It is easy to notice that we cannot state that **some** (P) (W) = **some** (P) (W'∩W'') =1 iff P∩W ≠ ϕ, where P=the set of Portugal players, and W'=the set of individuals who have won Euro 2016, and W''= the set of individuals who have left the team. The reason is that when W=(W'∩W'') it is *not* the case that **some** (P) (W'∩W'') = **some** (P) (W''∩W') =1 iff P∩W ≠ ϕ. The equivalence, **some** (P) (W'∩W'') = **some** (P) (W''∩W') =1, does not hold, and the commutative law for the intersection between W' and W'' does not apply here, contrary to the basic principles of set theory. This renders the verb phrase 'have won Euro 2016 and left the team' non-intersective, thereby eliminating the concurrent-situation reading. Therefore, the sentence in (17) has a collective split-situation reading which can be paraphrased in (18).

(18) *Some of the Portugal players have won Euro 2016 and then they have left the team.*

Clearly, (18) does not lead to oddness just as (17) does not, in that it basically states that there's a subset of Portugal players who won in the recent past and there's a subset of Portugal players who then left the team. Therefore, the crucial difference hinges on there being a collective *and* concurrent-situation reading in (4) but not in (17).

## 4. Conclusion

This paper suggests that the inferences within the *main effects* of a sentence (that is, within the truth-conditions) have something to do with the way predicates are restricted via conjunction. Most crucially, an SI-like effect may be generated through a *connected* sequence of propositions when entailed by a sentence whose SI is what is supposed to be determined. This may require further investigation, but this is beyond the scope of this paper and hence we leave this open.

## References

Anvari, A. (2018). Logical integrity. *Semantics and Linguistic Theory*, 28, 711–726.

Asher, N. (2012). Implicatures and discourse structure. *Lingua*, 132, 13-28.

Bassi, I., Del Pinal, G., & Sauerland, U. (2021). Presuppositional exhaustification. *Semantics and Pragmatics*, 14, 1-48.

Bott, L., & Noveck, I. A. (2004). Some utterances are underinformative: The onset and time course of scalar inferences. *Journal of Memory and Language*, 51(3), 437–457.

Burton-Roberts N. (2010). Cancellation and intention. In B. Soria & E. Romero (Eds.), *Explicit communication: Robyn Carston's pragmatics* (pp. 138-155). Palgrave Macmillan.

Chierchia, G., Fox, D., & Spector, B. (2011). The grammatical view of scalar implicatures and the relationship between semantics and pragmatics. In P. Portner, Claudia C. Maienborn & K. von Heusinger (Eds.), *Handbook of semantics* (pp. 2297–2332). Mouton de Gruyter.

Chierchia, G. (2013). *Logic in grammar: Polarity, free choice, and intervention*. Oxford University Press.

Del Pinal, G. (2021). Oddness, modularity, and exhaustification. *Natural Language Semantics*, 29, 115–158.

Enguehard, É., & Chemla, E. (2021). Connectedness as a constraint on exhaustification. *Linguistics and Philosophy*, 44, 79–112.

Fox, D. (2007). Free choice and the theory of scalar implicatures. In U. Sauerland & P. Stateva (Eds.), *Presupposition and implicature in compositional semantics* (pp. 71–120). Palgrave Macmillan.

Fox, D., & Katzir, R. (2011). On the characterization of alternatives. *Natural Language Semantics*, 19, 87–107.

Geurts, B. (2009). Scalar implicature and local pragmatics. *Mind and Language*, 24(1), 51-79.

Grice, H. Paul. (1975). Logic and conversation. In P. Cole & Jerry L. Morgan (Eds.), *Syntax and Semantics* (pp. 41–58). Academic Press.

Heim, I. (1991). On the projection problem for presuppositions. In S. Davis (Ed.), *Pragmatics: A reader* (pp. 397–405). Oxford University Press.

Horn, L. (1972). *On the semantic properties of the logical operators in English*. [Doctoral thesis, University of California]. Los Angeles.

Horn, L. (1989). *A natural history of negation*. Chicago University Press.

Horn, L. (2000). From IF to IFF: Conditional perfection as pragmatic strengthening. *Journal of Pragmatics*, 32, 289–326.

Katzir, R. (2007). Structurally-defined alternatives. *Linguistics and Philosophy*, 30, 669–690.

Keenan, Edward L., & Stavi, J. (1986). A semantic characterization of natural language determiners. *Linguistics and Philosophy*, 9(3), 253-326.

Keenan, Edward L. (2006). Quantifiers: Semantics. In K. Brown (Ed.), *Encyclopaedia of language and Linguistics* (2nd ed., vol. 10, pp. 302–308). Elsevier.

Levinson, Stephen C. (1983). *Pragmatics*. Cambridge University Press.

Levinson, Stephen C. (2000). *Presumptive meanings: The theory of generalized conversational implicatures*. MIT Press.

Magri, G. (2009). A Theory of individual-level predicates based on blind mandatory scalar implicatures. *Natural Language Semantics*, 17(3), 245-297.

Magri, G. (2011). Another argument for embedded scalar implicatures based on oddness in downward entailing environments. *Semantics & Pragmatics*, 4, 1-51.

Magri, G. (2016). Two puzzles raised by oddness in conjunction. *Journal of Semantics*, 33(1), 1-17.

Marty, P., & Romoli, J. (2022). Presupposed free choice and the theory of scalar implicatures. *Linguistics and Philosophy*, 45, 91-152.

Mayol, L., & Castroviejo, E. (2013). How to cancel an implicature. *Journal of Pragmatics*, 50(1), 84-104.

Pistoia-Reda, S., & Romoli, J. (2017). Oddness in conjunction. In S. Pistoia-Reda & F. Domaneschi (Eds.), *Linguistic and psycholinguistic approaches on implicatures and presuppositions* (pp. 55-71). Palgrave Macmillan.

Pistoia-Reda, S. (2017). Contextual blindness in implicature computation. *Natural Language Semantics*, 25, 109–124.

Poortman, Eva B. (2017). Concept typicality and the interpretation of plural predicate conjunction. In James A. Hampton & Y. Winter (Eds.), *Compositionality and concepts in Linguistics and Psychology* (pp. 139–162). Springer Nature.

Roberts, C. (1996). Information structure in discourse: Toward an integrated formal theory of pragmatics. *Ohio State University Working Papers in Linguistics*, 49, 91-136.

Sauerland, U. (2004). Scalar implicatures in complex sentences. *Linguistics and Philosophy*, 27, 367–91.

Schlenker, P. (2012). ''Maximize Presupposition'' and Gricean reasoning". *Natural Language Semantics*, 20, 391–429.

Spector, B., & Sudo, Y. (2017). Presupposed ignorance and exhaustification: How scalar implicatures and presuppositions interact. *Linguistics and Philosophy*, 40, 473–517.

Sperber, D., & Wilson, D. (1995). *Relevance: Communication and Cognition*. Blackwell.

Stalnaker, R. (1998). On the representation of context. *Journal of Logic, Language, and Information*, 7, 3–19.

Wilson, D., & Sperber, D. (2004). Relevance theory. In Laurence R. Horn & G. Ward (Eds.), *The handbook of pragmatics* (pp. 607–632). Blackwell.

Zhang, Z., Bergen, L., Paunov, A., Ryskin, R., & Gibson, E. (2023). Scalar implicature is sensitive to contextual alternatives. *Cognitive Science*, 47(2), e13238.